\documentclass[final,3p,twocolumn]{elsarticle}  
\usepackage{xcolor, soul, framed}

\colorlet{shadecolor}{yellow}
\usepackage{graphicx}
\usepackage{amsmath}
\usepackage{array}
\usepackage{mdwmath}
\usepackage{mdwtab}
\usepackage{eqparbox}
\usepackage{url}
\usepackage{hyperref}
\usepackage[utf8]{inputenc}
\usepackage{amssymb}
\usepackage{amsfonts}

\usepackage{algorithm}
\usepackage{algpseudocode}

\usepackage{comment}
\usepackage{todonotes}

\usepackage{caption}
\usepackage{subcaption}

\usepackage{placeins}

\begin{document}
\begin{frontmatter}

\title{Robust affine point matching via quadratic assignment on Grassmannians}

\author{Alexander Kolpakov}
\ead{kolpakov.alexander@gmail.com}

\author{Michael Werman}
\ead{michael.werman@mail.huji.ac.il}

\begin{abstract}
Robust Affine Matching with Grassmannians (RoAM) is a new algorithm to perform affine registration of point clouds. The algorithm is based on minimizing the Frobenius distance between two elements of the Grassmannian. For this purpose, an indefinite relaxation of the Quadratic Assignment Problem (QAP) is used, and several approaches to affine feature matching are studied and compared. Experiments demonstrate that RoAM is more robust to noise and point discrepancy than previous methods.
\end{abstract}

\end{frontmatter}
\section{Introduction}

\begin{figure}
    \centering
    \begin{minipage}{0.75\columnwidth}
        \centering
        \includegraphics[scale=0.18]{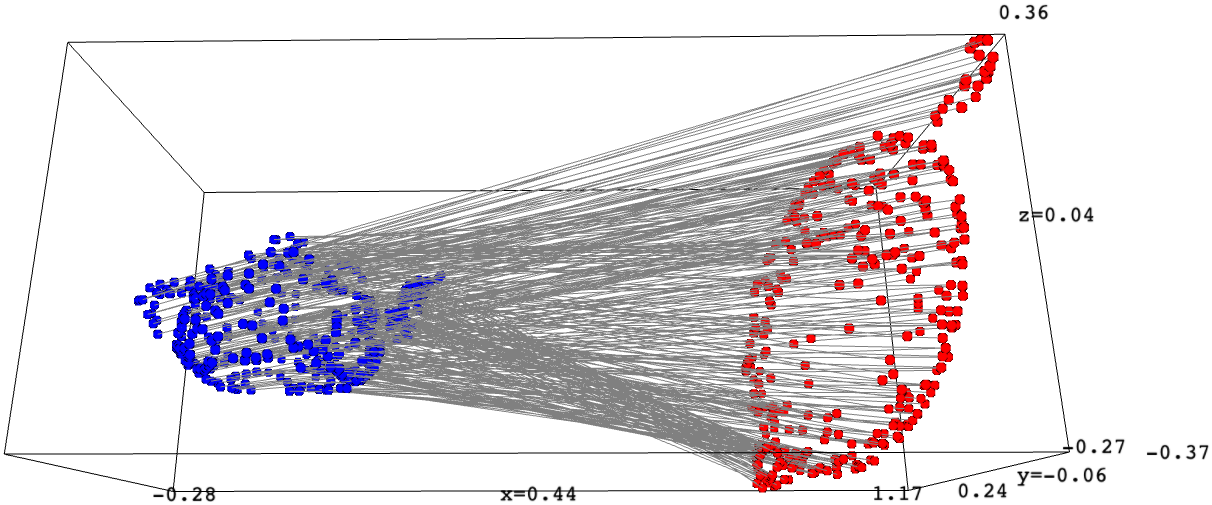}
    \end{minipage}
    ~
    \begin{minipage}{0.75\columnwidth}
        \vspace{0.25in}
    \end{minipage}
    ~
    \begin{minipage}{0.75\columnwidth}
        \centering
        \includegraphics[scale=0.18]{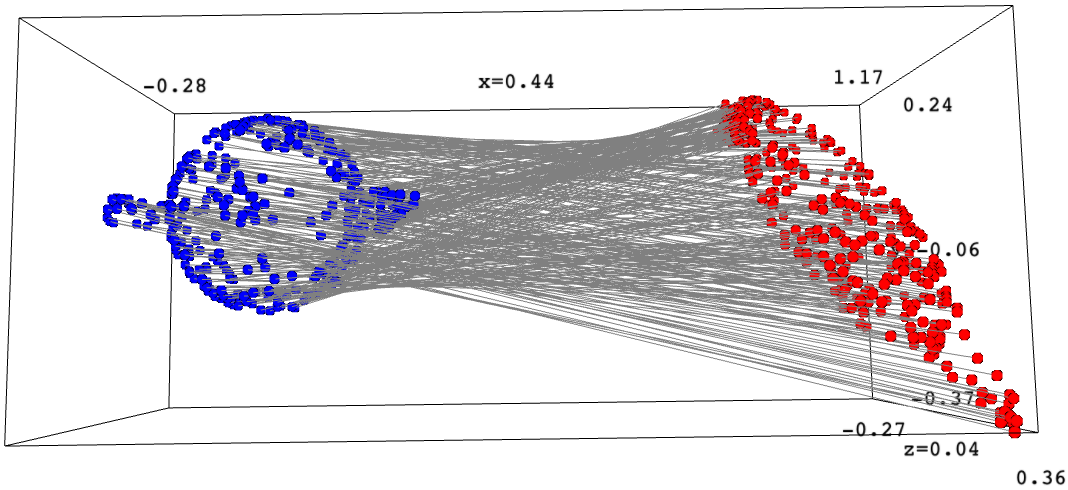}
    \end{minipage}
\caption{ ``Teapot'' cloud: the initial point cloud $X$ (blue)  and its image $Y$ (red) under a linear transformation $L$ with $\mathrm{cond}\,L = 4.0$. The feature matching is shown: side view (top) and top view (bottom).}\label{fig:teapot_features}
\end{figure}

\begin{figure}[b]

        \centering
        \includegraphics[scale = 0.15]{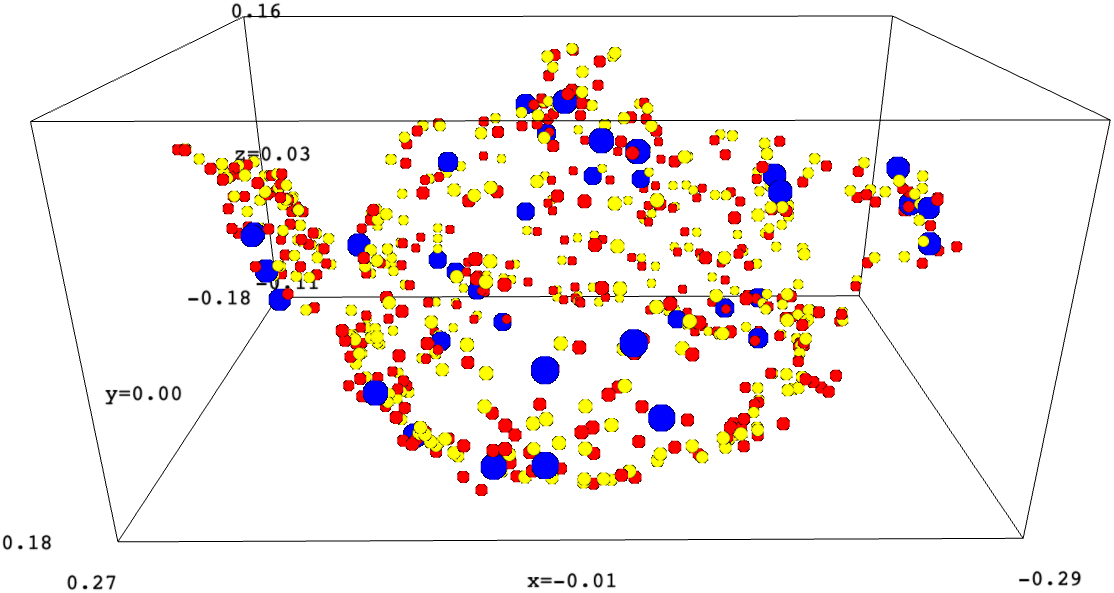}
        \caption{ ``Teapot'' cloud: specimen $X$ (yellow) and preimage $L^{-1}_0\, Y$ (red). Points of $X$ having no correspondence in $Y$ are marked (blue).}
        \label{fig:teapot_roam}
\end{figure}

 \begin{figure}
        \centering
        \includegraphics[scale = 0.14]{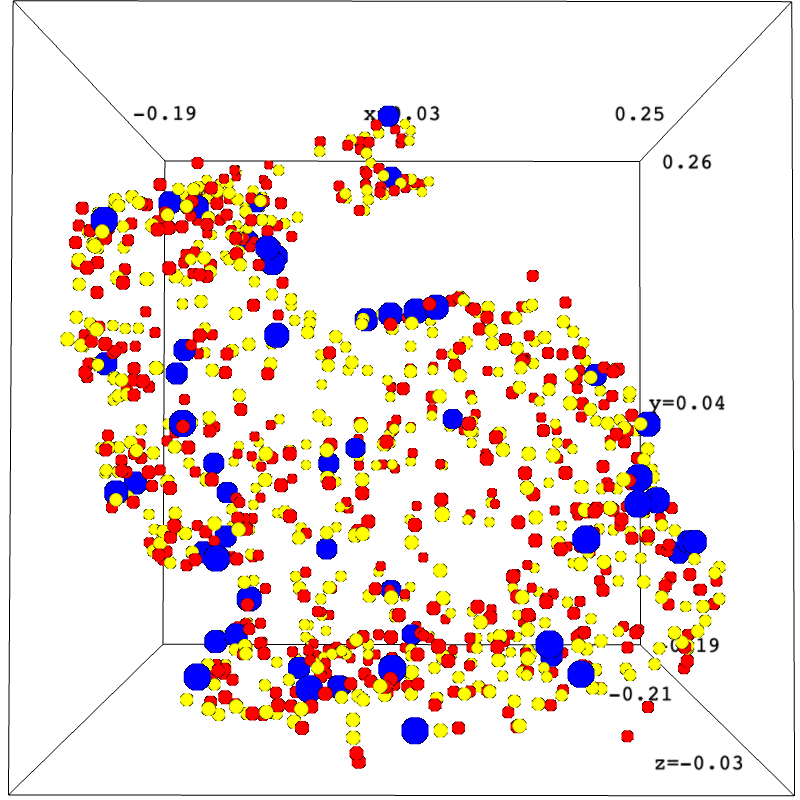}
        \caption{ ``Bunny'' cloud: specimen $X$ (yellow) and preimage $L^{-1}_0\, Y$ (red). Points of $X$ having no correspondence in $Y$ are marked (blue).}
        \label{fig:bunny_roam}
    \end{figure}

    \begin{figure}
        \centering
        \includegraphics[scale = 0.14]{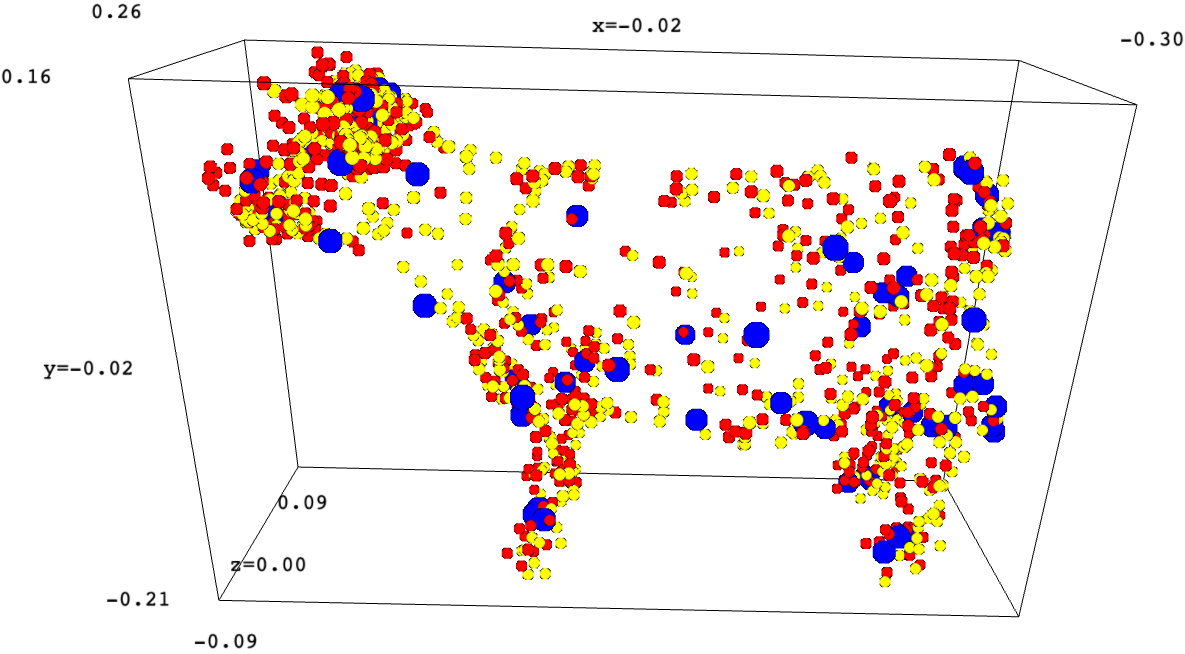}
        \caption{ ``Cow'' cloud: specimen $X$ (yellow) and preimage $L^{-1}_0\, Y$ (red). Points of $X$ having no correspondence in $Y$ are marked (blue).}
        \label{fig:cow_roam}
    \end{figure}

Affine registration has a long history in computer vision literature, and extensive work was carried out for affine registration in two and three dimensions, \cite{barrios2022pso,tang2013medical,121791,huang2021comprehensive,1641009}.  In \cite{10.1007/978-3-540-88693-8_19}, two interesting matching problems are formulated and solved based on affine registration  in higher dimensions; stereo correspondence under motion and image set~matching. 

This paper is  related to 
\cite{GrassGraph} which proposed a two-stage procedure  to recover correspondences between  unlabelled, affinely transformed point sets. 
The first step finds the affine transformation, and the second  is the matching between the point sets.
We adopt their first step of using the Grassmannian,  which is invariant under linear (affine) transformations of the point sets. 

As noted in \cite{Lyzinski_2016}, it is provably better, in certain cases, to use an {\it indefinite relaxation} of the Quadratic Assignment Problem (QAP), which we use to find the matching, while the common convex relaxation almost always fails. 

Based on these theoretical findings, we propose a relaxed quadratic programming method to identify the correct permutation (match) between the sets of points. The experimental outcomes  support  our notion that the new algorithm is more resistant to noise and point discrepancies than the approach presented in~\cite{GrassGraph}.

\section{Algorithm RoAM} 

The algorithm is based on two main ideas, first  affine invariant representations, Grassmanians, are computed for the point clouds and then a relaxation of a hard computational problem (QAP) is used to find the matching between the representative Grassmanians.

Let $X, Y \subset \mathbb{R}^{d \times n}$ be two $n$  point clouds in $\mathbb{R}^d$, such that $Y = L\,X\,S + t$ for a linear transformation $L \in \mathrm{GL}(d)$, a matrix $S$ from the group of  permutation matrices $\mathrm{Sym}(n)$, and a translation vector $t\in \mathbb{R}^d$. Let us assume that $t=0$ by translating both point clouds to have barycenters, an affine invariant, at $0$. Furthermore, we assume, generically, that $\mathrm{rank}\, X = d$, which implies the same rank for $Y$.

Let $\mathrm{Gr}(k, n)$ be the Grassmannian of $k$--dimensional planes in $\mathbb{R}^n$. We can view the point cloud $X$ as a map $X: \mathbb{R}^n \rightarrow \mathbb{R}^d$  defined by $X(v) = X \, v$, for $v \in \mathbb{R}^n$. Then $\ker X$ is an element of $\mathrm{G} = \mathrm{Gr}(n-d, n) \cong \mathrm{Gr}(d, n)$.  Moreover, for any $L \in \mathrm{GL}(d)$ we have $\ker LX = \ker X$. Thus, $X$ and all of its non--degenerate images define the same element of $\mathrm{G}$. 
Let $P_X$ be the orthogonal projection of $\mathbb{R}^n$ onto $\ker X^\perp$. The orthogonal projection $P_X$, which uniquely defines $\ker X$, will be used  to  represent an element of $\mathrm{G}$. Note that $P_X$ can be  computed from the (reduced) singular value decomposition (SVD) of $X$,

\begin{equation}
  X = U\, \cdot \Sigma \, \cdot V
    \Rightarrow
  P_X = V^T \, V,  
\end{equation}
where $U \in \mathbb{R}^{d\times d}$, $\Sigma \in \mathbb{R}^d$ is the diagonal of non--zero singular values, and $V \in \mathbb{R}^{d \times n}$.

We also have $P_Y = S^T\, P_X\, S$. 
The matrix $S$ is not unique if there is a linear transformation $T \in \mathrm{GL}(d)$ such that $T\, X = X\, S$. Generically this does not happen, which we formulate as ``$X$ has no symmetries''. 

Thus, we can reformulate the initial problem as
\begin{equation}
    \| P_Y - S^T\, P_X\, S \|_F \rightarrow \min_{S \in \mathrm{Sym}(n)},  
\end{equation}
which is equivalent to 
\begin{equation}
    \mathrm{tr}(P_Y\,S^T\,P_X\,S) \rightarrow \max_{S \in \mathrm{Sym}(n)}.
\end{equation}
This is the Quadratic Assignment Problem (QAP) \cite{koopmans1957assignment} and is NP--complete.

A generic picture of the ``Teapot'' point cloud $X$ and its image $Y$ is shown in Figure~\ref{fig:teapot_features}. The feature matching under the linear map $L$ between the points is marked. However, the points of $Y$ may be randomly permuted, and determining the right matching $S$ is  necessary to recover the linear map $L$. 

Rather than directly optimizing over the permutation matrices,  we  relax the constraint set to the convex hull of $S$, the set of doubly stochastic matrices (also known as the Birkhoff polytope),
\begin{equation}
    \mathcal{D}=\{S\in [0,1]^{n\times n} ~\big{|}~ S 1_n=1_n, S^t 1_n=1_n\},
\end{equation}
non--negative square matrices whose rows and columns all sum to one.

Relaxing the feasible region allows one to employ the tools of continuous optimization to search for local optima, as this is not a convex problem  different initializations can give different results.

In the ideal case when $P_Y = S^T\, P_X \, S$ we have that
\begin{equation}
    \max_{S \in \mathrm{Sym}(n)}\mathrm{tr}\,P_Y\,S^T\,P_X\,S = \mathrm{tr}\, P^2_Y = \mathrm{tr}\, P_Y = d.
\end{equation}
The penultimate equality holds because $P^2 = P$ for any projection operator $P$. 
    
Thus, given a matrix  $S$ produced by rQAP, we may assign a weight $w(S)$ to it that measures the proximity of $S$ to $S_*$. For example, 
\begin{equation}
    w(S) = \exp\left(\, -\, C \cdot (\mathrm{tr}(P_Y S^T P_X S) - d)^2 \,\right),
\end{equation}
where $C > 0$ is some large constant. The   requirement  is that $w(S) \approx 1$ for  permutations $S$ that are Hamming--close to $S_*$ and $w(S) \approx 0$ for all others.

The idea is to run a sufficiently large number, $N$, of trials of rQAP, with different initializations obtaining the solutions $S_1$, $\ldots$, $S_N$, and then project
\begin{equation}
    \pi: \sum^N_{i=1} w(S_i)\, S_i \longmapsto S_0 \in \mathrm{Sym}(n).
\end{equation}
The projection $\pi$ to the nearest permutation can be realized e.g. by the Hungarian Algorithm.

Another approach would be to take such an $S_*$ among $S_1, \ldots, S_N$ that provides the best match between $P_X$ and $P_Y$, that is 
$ S_*=\textrm{argmin}_{S_i} \| P_Y - S_i^T P_X S_i \|_F$. 
The results of \cite{Lyzinski_2016} suggest that this indefinite relaxation of QAP proposed in \cite{Vogelstein_2015} finds the optimal solution, $S_*$, with high probability. We shall refer to the algorithm proposed in \cite{Lyzinski_2016} as rQAP (relaxed Quadratic Assignment Problem).

We compare the best match and weighted sum approaches in  
section ~\ref{ffm} below. 

Once $S$ is recovered,  we recover $L$ as
\begin{equation}
    L = Y\,S^T\,X^T\,(X\,X^T)^{-1},
\end{equation}
since $L$  solves the least squares problem 
\begin{equation}
    \min_{L \in \mathrm{GL}(d)} \| L\, X - Y\, S^T  \|_F.
\end{equation}

\begin{algorithm}
\caption{RoAM: Robust Affine Matching with Grassmannians}\label{alg}
\begin{algorithmic}
\Require{$X, Y \subset \mathbb{R}^{d\times n}$} 
\Ensure{ $L_0$ and $S_0$ so that  $Y \approx L_0\,X\,S_0$}
\\

    \State $X \leftarrow \overline{X}$, $Y\leftarrow \overline{Y}$ 
    \Comment{centering}
    \State $P_X \leftarrow  \ker X$, 
    $P_Y \leftarrow  \ker Y$
 \For{$i \gets 1$ to $N$}  
     $S_i \leftarrow 
    \max_{S \in \mathcal{D}} \mathrm{tr}(P_Y\,S^T\,P_X\,S)$ \Comment{rQAP}
\EndFor  
    \State  $S_0 \leftarrow \pi \left( \sum^N_{i=1} w(S_i)\, S_i  \right)$
    \Comment{Hungarian Algorithm}
    \State$L_0 := Y\,S^T_0\,X^T\,(X\,X^T)^{-1}$
\end{algorithmic}
\end{algorithm}

\subsection{Correctness}

Let $S$ be the feature matching matrix produced by RoAM. Let $\mathbb{P}(S = S_*) = 1 - \alpha$ for some constant $0 \leq \alpha < 1$, and $\mathbb{P}(S \neq S_*) = \alpha$. In the setting of best matching for random Bernoulli graphs $\alpha = 0$  holds \cite{Vogelstein_2015}. 

Let $w: \mathcal{D} \rightarrow [0, 1]$ be a weight function on the set of doubly stochastic matrices $\mathcal{D} \supset \mathrm{Sym}(n)$ such that $w(S_*) = 1$ while $w(S') \leq \varepsilon$. Let $S_w = w(S) \, S$ be the weighted version of a permutation $S\in \mathrm{Sym}(n)$. Then
\begin{align}
&(1-\alpha) S_* \leq \mathbb{E}\, S_w  \nonumber \\
&=(1 - \alpha)\, w(S_*)\, S_* +
 \sum_{S'\in \mathrm{Sym}(n),\; S' \neq S_*} \mathbb{P}(S = S')\, w(S)\, S \nonumber \\
 &\leq (1 - \alpha) S_* + \varepsilon \alpha \mathbf{1}^T \mathbf{1}
\end{align}

Thus, once $\varepsilon < \frac{1-\alpha}{\alpha}$ we have that $\pi(\mathbb{E}\, S_w) = S_*$. However, we need to guarantee that the same equality holds once we replace $\mathbb{E}\,S_w$ by the sample average $\overline{S}_w$ for a large enough sample $\{S_i\}^N_{i=1}$ of permutation matrices produced by RoAM. 


Let the sample $\{S_i\}^N_{i=1}$ contain $m \leq N$ matrices $S_i = S_*$ and $N - m \geq 0$ matrices $S_i \neq S_*$. Then we have $\alpha = 1 - \frac{m}{N}$, and thus $\pi(\overline{S}_w) = S_*$ for any $\varepsilon < \frac{m}{N-m}$. 


The Cauchy--Bunyakovsky inequality implies that
\begin{align}
    &| \mathrm{tr}(P_Y S^T P_X S) | \leq \| P_Y S^T \|_F \, \| P_X S\|_F \nonumber \\
    &= \| P_Y \|_F \| P_X \|_F = d,
\end{align}
and thus, we need to choose a sufficiently large constant 
\begin{equation}
    C \geq \frac{\log\left(1/\varepsilon\right)}{4 d^2}
\end{equation}
in the equality
\begin{equation}
    w(S) = \exp\left(\, - C \cdot (\mathrm{tr}(P_Y S^t P_X S) - d)^2 \,\right).
\end{equation}

\subsection{Robustness to noise}

As follows from Wedin's theorem \cite{Wedin:1972} (see also Theorem~2.5 and Corollary 2.6 in \cite{Stewart:1973}), given a point clouds $Y \subset \mathbb{R}^{d \times n}$ a small perturbation of the vectors in $Y$ results only in a small perturbation of the projection $P_Y$ as the subspace $\ker Y^\perp$, corresponds to the $d$ largest singular values of  $Y$. This means that $\ker Y$ is only slightly perturbed in the sense that if $\widetilde{Y} = Y + N$, where $N$ represents noise, then $\| P_Y - P_{\widetilde{Y}} \|_2 = O(\|N\|_2)$. 

\subsection{Point discrepancy}

Let $X \subset \mathbb{R}^{d \times m}$ and $Y \subset \mathbb{R}^{d \times n}$, where we have $m \leq n$. Assume that there exist two matrices $L \in \mathrm{GL}(d)$ and $S \in \{0,1\}^{n \times m}$ such that $Y' = Y\, S = L\, X$. 

Then we replace the projection matrix $P_X \in \mathbb{R}^{m \times m}$ with $P_X \leftarrow P_X \oplus \mathbf{0}_{n-m}$ and apply RoAM to the new pair $P_X, P_Y \in \mathbb{R}^{n \times n}$. Once an output $S_0$ is produced, we pick only the first $m$ columns to produce a matching between the points of $X$ and $Y$. Indeed, matching any of the extra $n-m$ rows/columns of $P_Y$ to a zero row/column in $P_X$ will be heavily penalized, and thus rQAP will target the closest possible match for the first $m$ rows/columns of $P_X$ and some $m$ rows/columns of $P_Y$. Based on our experiments, this approach works well with missing/added points in $X$ and $Y$, even for relatively large discrepancies. 

\subsection{Finding feature matching}
\label{ffm}
\begin{figure}
    \centering
    
    \includegraphics[scale=0.15]{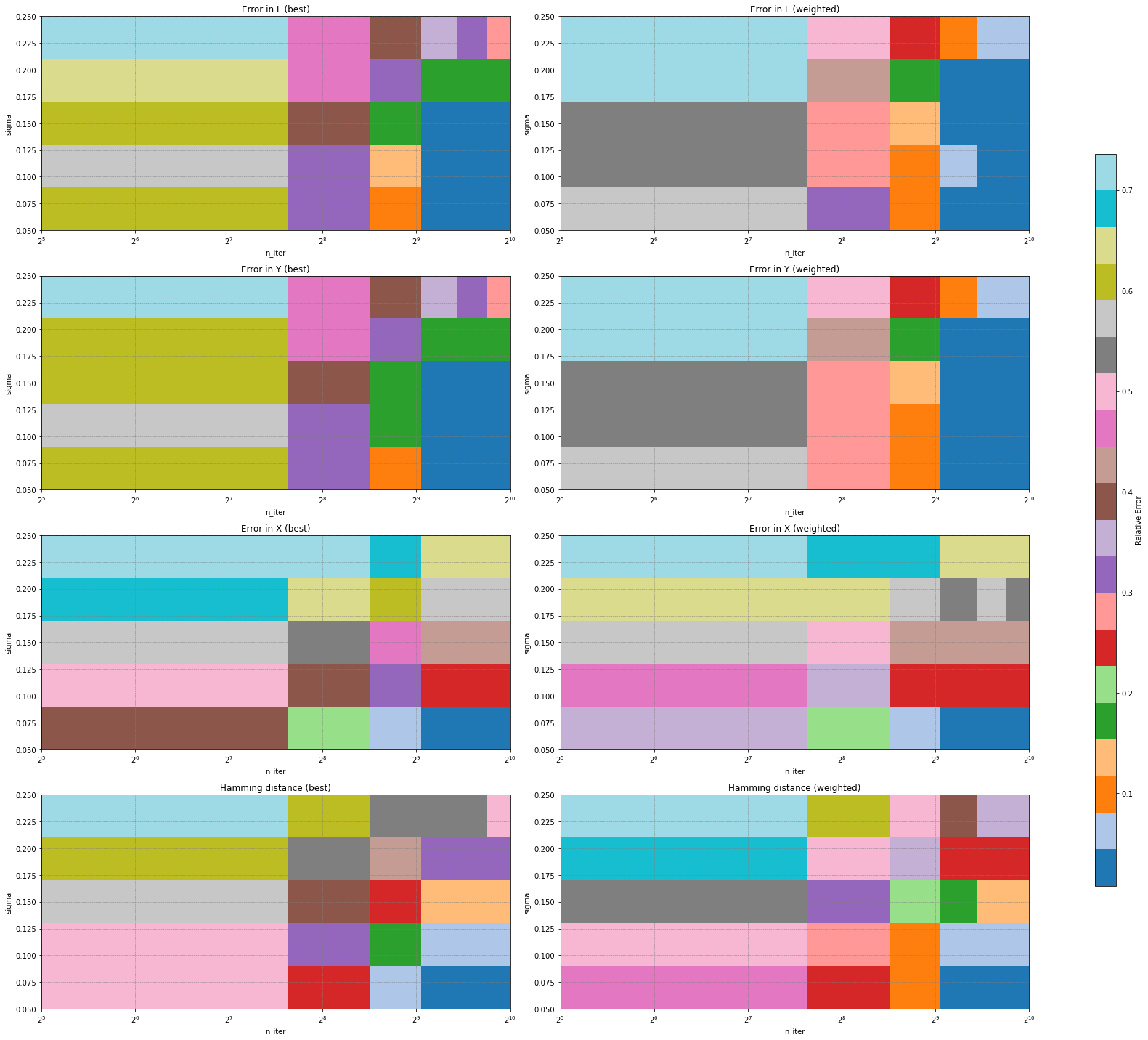}
    \caption{Best matching (left) vs. weighted sum (right) approach for restoring the feature correspondence. The relative errors $\delta_L$, $\delta_Y$, and $\delta_X$ are shown (top to bottom). The relative Hamming distance between the recovered and ground truth feature matching is measured (bottom row). The horizontal axis of each heatmap is the number of iterations $N$, the vertical axis is $\sigma$ for the multiplicative Gaussian noise $\mathcal{N}(1, \sigma^2)$ affecting $Y$.}\label{best-vs-weighted}
\end{figure}

In Figure~\ref{best-vs-weighted} we compare the relative errors $\delta_L$, $\delta_Y$ and $\delta_X$ as defined above for the two approaches: 

\begin{itemize}
    \item[1.] Taking the best match $S_0$ solving $\| P_Y - S^T P_X S \| \rightarrow \min_{S \in \{S_1, \ldots, S_N\}}$;
    \item[2.] Taking the weighted sum $S_0 = \pi\left( \sum^N_{i=1} w(S_i)\,S_i \right)$.
\end{itemize}

The relative Hamming distance $$\delta_H = \frac{1}{2n} \| S - S_0 \|^2_F$$ between the recovered $S_0$ and ground truth $S$ feature matching of $n$ points is measured: for this purpose we assume that $X$ and $Y$ have no point discrepancy.

The initial point cloud $X$ is a random point cloud of $n$ points distributed uniformly in $[0,1]^d$, and the image point cloud $Y = L\,X\, S$ is obtained by applying a random linear map $L$ with given condition number, and a random permutation matrix $S \in \mathrm{Sym}(n)$. We also apply Gaussian multiplicative noise $\mathcal{N}(1, \sigma^2)$ to each entry of $Y$. 

In Figure~\ref{best-vs-weighted} we have $d=3$, $n=100$, and $\mathrm{cond}\,L = 3.0$. The noise magnitude (determined by $\sigma$) and the number of iterations $N$ in the feature matching search vary: namely $\sigma \in [0.05, 0.25]$ (with a step of $0.05$) and $N = 2^k$ for $k = 5$ up to $k=10$ with unit step. For each pair of parameters $(\sigma, N)$ we perform $100$ tests and report the average results. 

As evidenced in Figure~\ref{best-vs-weighted} neither approach shows acceptable accuracy for $N \leq 2^8$. For $N \geq 2^9$ the best match approach shows equally good or better accuracy while the noise is relatively small ($0.0 < \sigma < 0.15$), while the weighted sum approach appears to be more robust for noise of larger magnitudes ($0.15 < \sigma < 0.25$).

\section{Experiments}
The following quantities were computed for  each experiment:
\begin{itemize}
    \item the error in recovering the affine transformation $L$,  $\delta_L = \| L - L_0 \|_2 / \| L \|_2$;
    \item the error in matching the preimages defined as $\delta_X = \| [L^{-1}_0\, Y\, S^T]_X - X \|_2 / \| X \|_2$, where $[A]_B$ is the matrix formed from the columns of $A = L^{-1}_0\, Y\, S^T \in \mathbb{R}^{d \times n}$ indexed with the same indexes as the columns of $B = X \in \mathbb{R}^{d \times m}$ inside $X' \in \mathbb{R}^{d \times n}$;
    \item the error in matching the images,   $\delta_Y = \| L\, X - L_0\, X \|_2 / \| L\,X \|_2$.
\end{itemize}

\subsection{Random point clouds}

In our numerical experiments, RoAM was applied to pairs of synthetic point clouds with the following specifications: one specimen point cloud $X \subset \mathbb{R}^{3 \times m}$ and another point cloud $Y \subset \mathbb{R}^{3 \times n}$ such that 
\begin{itemize}
    \item $X$ is a random subset of $X' \subset \mathbb{R}^{3 \times n}$ with $m = \lfloor \lambda \,n \rfloor$ points with $\lambda \in (0, 1]$ being the level of similarity (with $\lambda = 1$ if all of $X$ coincides with $X'$); 
    \item $Y = Y' \odot N$, with $Y' = L\, X'\, S$, $L \in \mathrm{GL}(3)$ a random linear map such that $L = P\,O$ with $P$ a positive definite matrix having condition number $c \in [1, +\infty)$ and $O$ a uniformly distributed orthogonal matrix; $S \in \mathrm{Sym}(n)$ a random permutation; and $N$,  multiplicative noise such that $N_{ij} \in \mathcal{N}(1, \sigma^2)$, $i=1,\ldots,3$, $j=1,\ldots,n$, are i.i.d. Gaussian variables. Here $\odot$ is the Hadamard (entry--wise) matrix product. 
\end{itemize}

\subsection{Caerbannog point clouds}

We used the (unoccluded) Caerbannog clouds \cite{caerbannog:2020} as the primary source of specimens $X$, while the corresponding $Y$ clouds were synthesized.

The following values for $\sigma$ (noise level) and $\lambda$ (similarity level to create point discrepancy) were used: 
\begin{itemize}
    \item $\sigma \in \{0.0, 0.01, 0.05, 0.1, 0.15, 0.2\}$,
    \item $\lambda \in \{1.0, 0.95, 0.90, 0.85, 0.8, 0.7, 0.6, 0.5\}$.
\end{itemize}

It appears to us that there is no need to increment $\sigma$ uniformly as we need to test only for reasonably small levels of noise (to observe that the algorithm works robustly) and for relatively high ones (to understand the limits of the algorithm's applicability). It also appears that we do not need to use lower levels of similarity than $0.5$ as  more than half of the points missing constitutes a fairly large discrepancy. As in the case of noise, we prefer to vary the discrepancy non--uniformly. 

For each value of the noise  $\sigma$ and point discrepancy  $\lambda$, a batch of $10$ tests was run and the mean values of $\delta_L$, $\delta_Y$, and $\delta_X$ were computed. Every run of RoAM used $N=2^{10}$ calls to rQAP. 

The Python code to perform the tests and the respective output stored  is accessible on GitHub \cite{github:2023}. The code requires SageMath \cite{sagemath} and runs in the Jupiter environment. The output  is saved in \texttt{csv} format and can be accessed without  running the code.  

For different levels of noise and occlusion, we got $69.43$ seconds on average per batch of $100$ trials for the ``choose best'' approach and $57.07$ seconds on average per batch of $100$ trials for the ``weighted sum'' approach (see \cite{github:2023}).  Thus, a single run of RoAM took approximately between $0.57$ to $0.69$ seconds on a MacBook Pro (M1, 2020). 

Below we provide a graphical interpretation of the results, where for each value of $\sigma$, the values of $\delta_*$, $* \in \{L, X, Y\}$ are provided for different values of $\lambda$. The higher the  point discrepancy (i.e. the lower $\lambda$), the longer  the horizontal bar over  the respective value of $\delta_*$, $* \in \{L, X, Y\}$. 

\begin{figure}[h]
\centering
\begin{minipage}{.47\columnwidth}
  \centering
  \includegraphics[width=\linewidth]{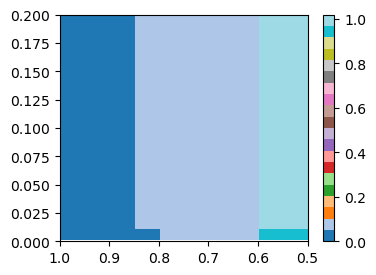}
  \captionof{figure}{``Teapot'' cloud: $\delta_L$ as a function of $\lambda$ (discrepancy level, horizontal) and $\sigma$ (noise level, vertical) }
        \label{fig:teapot_err_L}
\end{minipage}%
\hspace{0.075cm}
\begin{minipage}{.47\columnwidth}
  \centering
  \includegraphics[width=\linewidth]{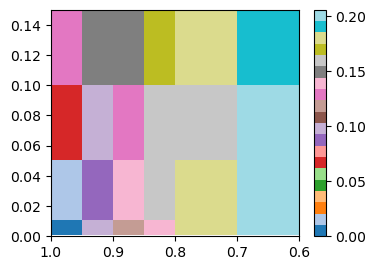}
  \captionof{figure}{``Teapot'' cloud: $\delta_X$ as a function of $\lambda$ (discrepancy level, horizontal) and $\sigma$ (noise level, vertical) }
  \label{fig:test1}
\end{minipage}
\end{figure}

\begin{figure}
\centering
\begin{minipage}{.47\columnwidth}
  \centering
  \includegraphics[width=\columnwidth]{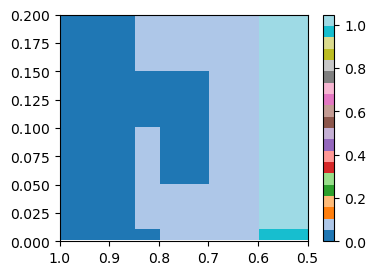}
  \caption{ ``Teapot'' cloud: $\delta_Y$ as a function of $\lambda$ (discrepancy level, horizontal) and $\sigma$ (noise level, vertical) }
        \label{fig:teapot_err_Y}      
\end{minipage}%
\hspace{0.075cm}
\begin{minipage}{.47\columnwidth}
  \centering
  \includegraphics[width=\linewidth]{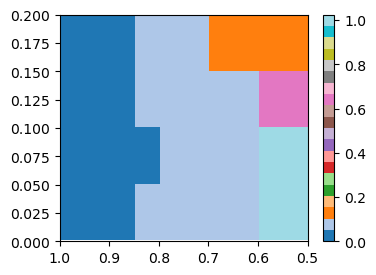}
  \caption{ ``Bunny'' cloud: $\delta_L$ as a function of $\lambda$ (discrepancy level, horizontal) and $\sigma$ (noise level, vertical) }
        \label{fig:bunny_err_L}  
    
\end{minipage}
\end{figure}

\begin{figure}
        \centering
        \begin{minipage}{.47\columnwidth}
        \includegraphics[scale=0.42]{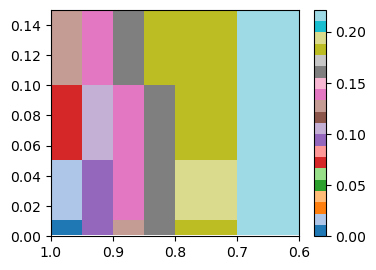}
        \caption{ ``Bunny'' cloud: $\delta_X$ as a function of $\lambda$ (discrepancy level, horizontal) and $\sigma$ (noise level, vertical) }
        \label{fig:bunny_err_X}
    \end{minipage}%
\hspace{0.075cm}
\begin{minipage}{.47\columnwidth}
        \centering
        \includegraphics[scale=0.42]{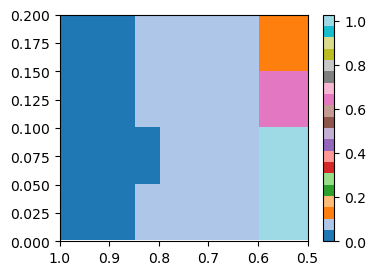}
        \caption{ ``Bunny'' cloud: $\delta_Y$ as a function of $\lambda$ (discrepancy level, horizontal) and $\sigma$ (noise level, vertical) }
        \label{fig:bunny_err_Y}
    \end{minipage}
    \end{figure}

\begin{figure}
    \begin{minipage}{.47\columnwidth}
        \centering
        \includegraphics[scale=0.42]{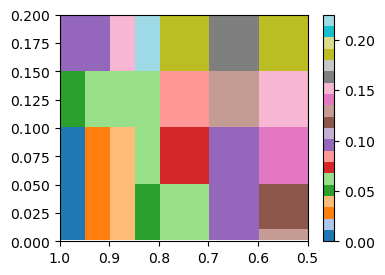}
        \caption{ ``Cow'' cloud: $\delta_L$ as a function of $\lambda$ (discrepancy level, horizontal) and $\sigma$ (noise level, vertical) }
        \label{fig:cow_err_L}
    \end{minipage}%
\hspace{0.075cm}
\begin{minipage}{.47\columnwidth}
        \centering
        \includegraphics[scale=0.42]{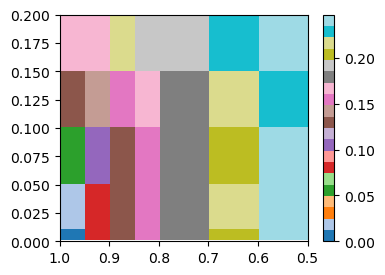}
        \caption{ ``Cow'' cloud: $\delta_X$ as a function of $\lambda$ (discrepancy level, horizontal) and $\sigma$ (noise level, vertical) }
        \label{fig:cow_err_X}
        \end{minipage}
    \end{figure}

In Figures \ref{fig:teapot_err_L} -- \ref{fig:cow_err_X}, we provide only values $\delta_* \leq 1.5$, and sometimes cannot avoid the visual merging and overlaps of some of the horizontal bars. The exact values of $\delta_*$, $*\in\{L,X,Y\}$ can be found in the  GitHub repository \cite{github:2023}. 

As part of the data available in \cite{github:2023}, we also measured the quantities
\begin{enumerate}
    \item $d_\sigma = \| Y' - Y \|_2 / \| Y \|_2$ (the relative noise  compared to the ground truth image of $X$ under $L$); and
    \item $d_\lambda = \| X' \setminus X \|_2 / \| X \|_2$ (the relative point discrepancy  compared to the initial specimen image).
\end{enumerate}

The \texttt{csv} files of the test data contain the following values in each line  in this order: $\sigma$, $\lambda$, $d_\sigma$, $d_\lambda$, $\delta_L$, $\delta_Y$, $\delta_X$. We are mostly interested in measuring $\delta_L$ and $\delta_Y$  to understand how well we can recover $L$ and how well the recovered linear map $L_0$ approximates the original image of $X$ under the action of $L$. The distance between the preimage of $Y$ under $L_0$ and the specimen image $X$ is used for comparison, and we use preimages for illustrative purposes (as the direct images under $L$ can be fairly deformed and thus less suitable for visualization). 

The linear transformation $L$ in all tests had condition number $\mathrm{cond}\,L = \| L \|_2\, \| L^{-1} \|_2$ equal to $3$. 

In Figures~\ref{fig:teapot_roam}, \ref{fig:bunny_roam} and \ref{fig:cow_roam}, we show the specimen cloud $X$ (yellow), the preimage $[L^{-1}\,Y\,S^T]_X$ (red), and also the point discrepancy consisting of points in $X'$ that do not belong to $X$ (blue). For all three point clouds we have $\sigma = 0.05$, $\lambda=0.90$, $\mathrm{cond}\,L = 3$. The number of trials in each case is $N = 2^{10}$. 


In Figure~\ref{fig:teapot_roam} we illustrate  a test with $X$ having $315$ points and $Y$ having $351$ points. The test parameters are $d_\sigma \approx 0.045$, $d_\lambda \approx 0.328$, with the output $\delta_L \approx 0.045$, $\delta_Y \approx 0.042$, and $\delta_X \approx 0.140$.


In Figure~\ref{fig:bunny_roam} we illustrate  a test with $X$ having $475$ points and $Y$ having $528$ points. The test parameters are $d_\sigma \approx 0.044$, $d_\lambda \approx 0.306$, with the output $\delta_L \approx 0.053$, $\delta_Y \approx 0.044$, and $\delta_X \approx 0.127$.


In Figure~\ref{fig:cow_roam} we illustrate a test with $X$ having $541$ points and $Y$ having $602$ points. The test parameters are $d_\sigma \approx 0.048$, $d_\lambda \approx 0.348$, with the output $\delta_L \approx 0.036$, $\delta_Y \approx 0.036$, and $\delta_X \approx 0.141$.

\subsection{Partial scans of the same object}

\begin{figure}
    \centering
\begin{subfigure}{.3\columnwidth}
    \centering
    \includegraphics[scale=0.25]{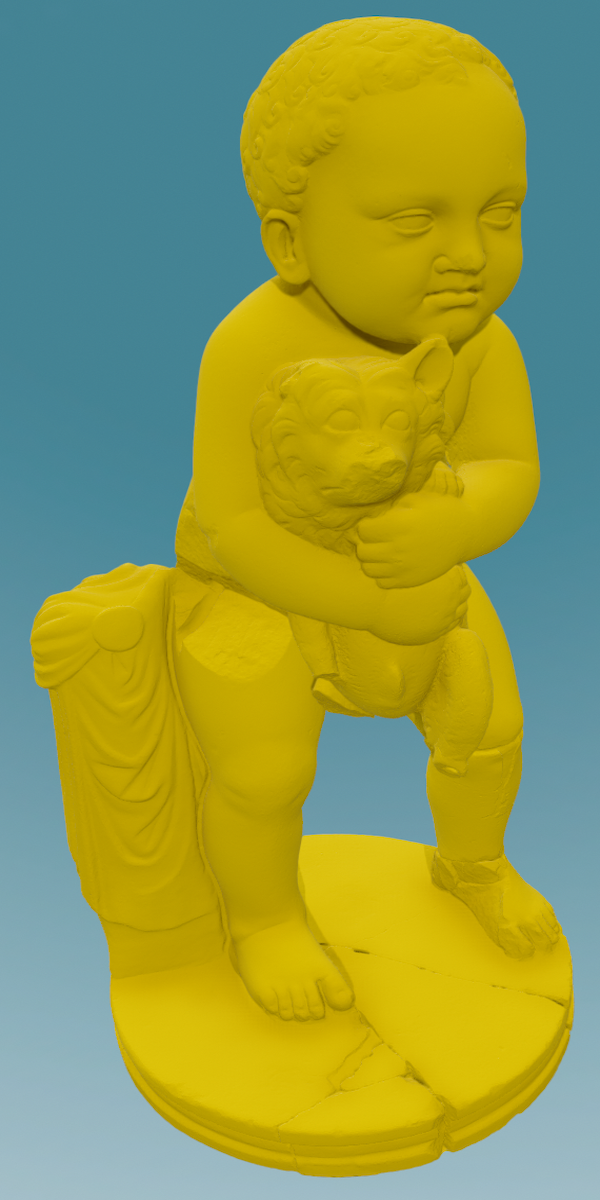}
    \end{subfigure}
    ~
\begin{subfigure}{.3\columnwidth}
    \centering
    \includegraphics[scale=0.25]{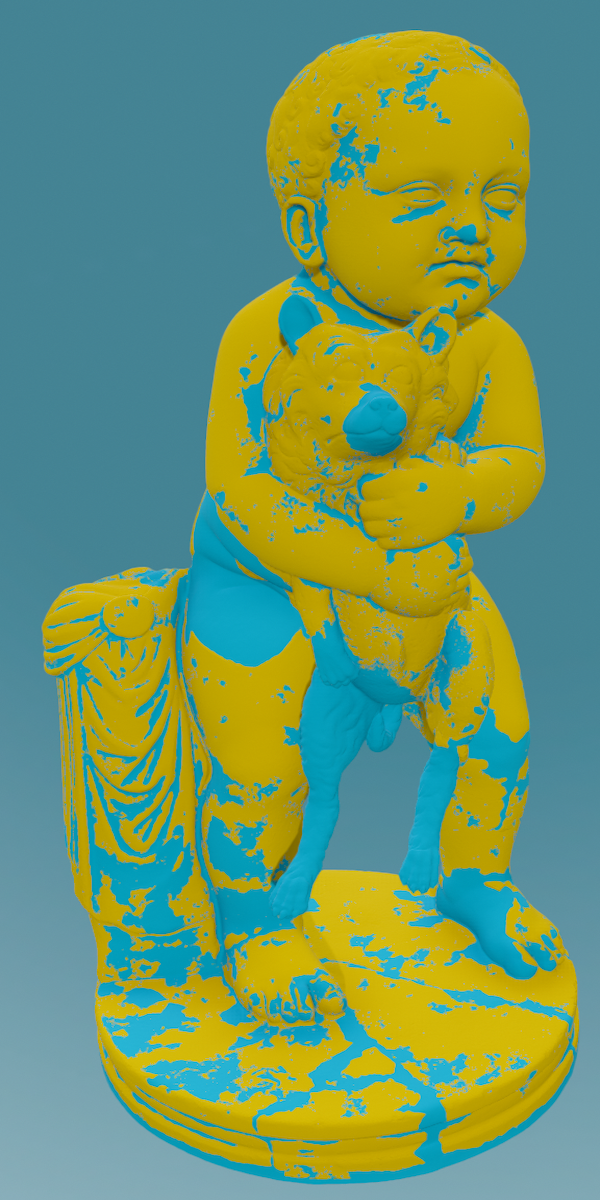}
    \end{subfigure}
    ~
\begin{subfigure}{.3\columnwidth}
    \centering
    \includegraphics[scale=0.25]{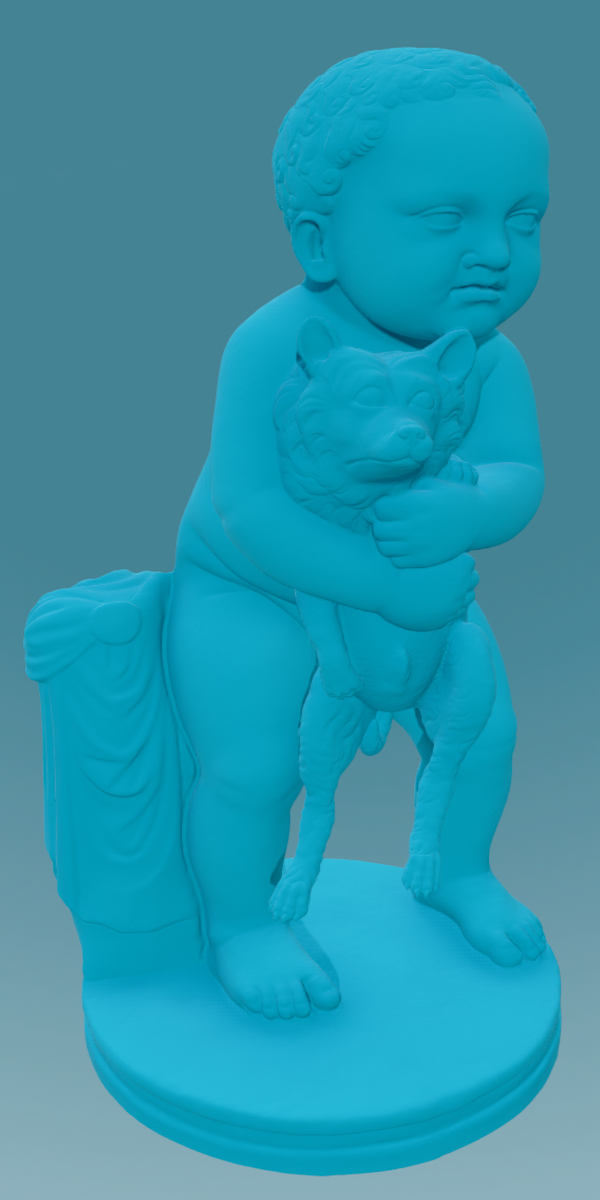}
    \end{subfigure}
    \caption{``Enfant au Chien'' before (left) and after (right) restoration (available from \cite{3-d-scans}). The ground truth of matching the sculptures (center).}\label{statue-broken-restored}
\end{figure}

We also use two point clouds sampled from two scanned images of the same sculpture before and after restoration. Namely, our specimen cloud $X$ is sampled from the mesh representing a scan of the damaged sculpture in Figure~\ref{statue-broken-restored} (left). The other cloud $Y$ is sampled from the mesh of a scan of the restored statue in Figure~\ref{statue-broken-restored} (right). More precisely, a random linear transformation with a given condition number is applied to the mesh, and only then $Y$ is sampled from it.

\begin{figure}
    \begin{subfigure}{.3\columnwidth}
    \centering
    \includegraphics[scale=0.25]{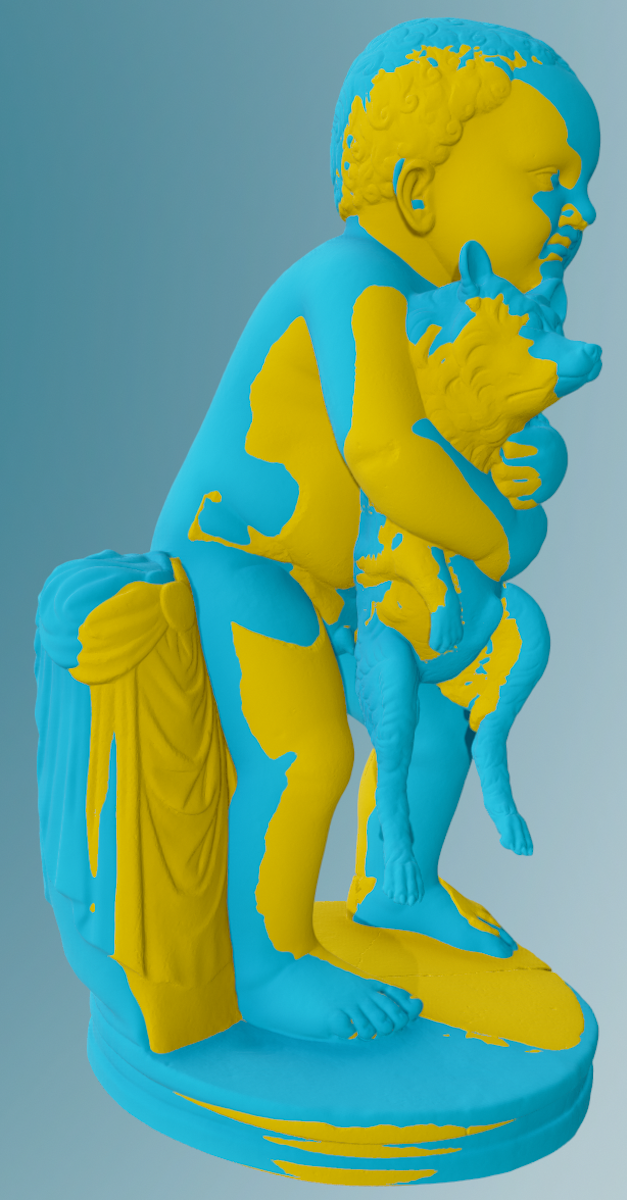}
    \end{subfigure}
    ~
\begin{subfigure}{.3\columnwidth}    \centering
    \includegraphics[scale=0.25]{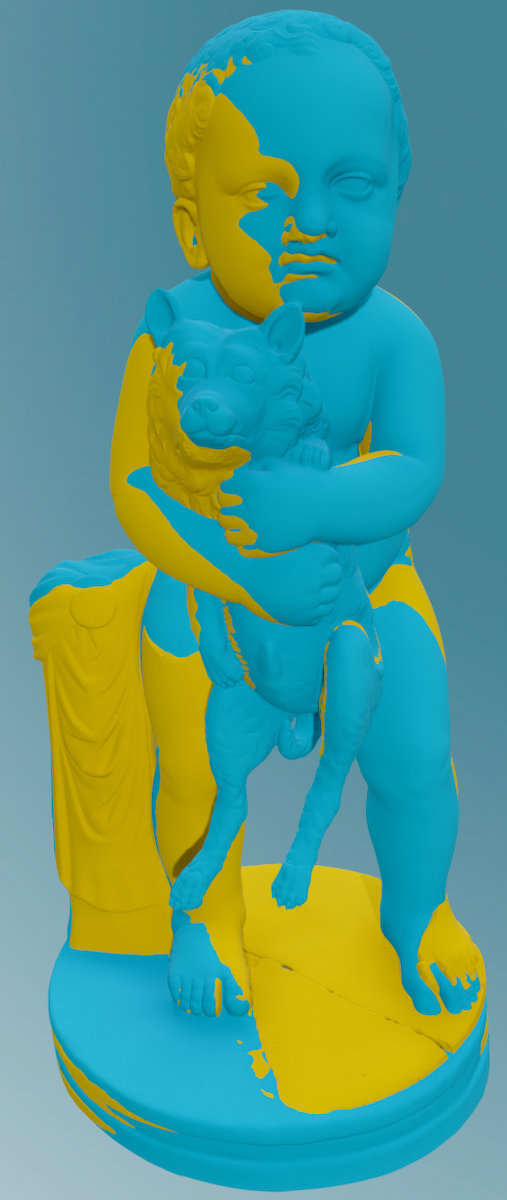}
    \end{subfigure}
    ~
\begin{subfigure}{.3\columnwidth}    \centering
    \includegraphics[scale=0.25]{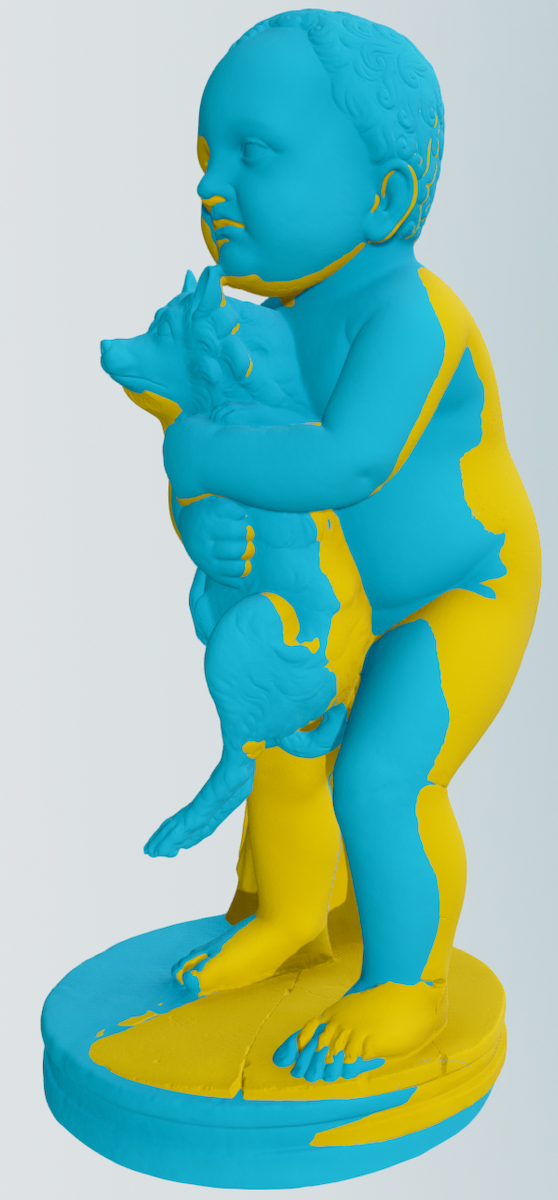}
        \end{subfigure}
    \caption{Testing RoAM: sample $600$ points from the broken statue mesh (source) for point cloud $X$, apply a random linear map $L$ with $\mathrm{cond}\,L = 3.0$ to the restored statue mesh (target) and sample another $600$ points from it for point cloud $Y$. Showing the  source and the image of the target under $L^{-1}_0\,L$. Frontal and side views.}\label{statue-600-600-3.0}
\end{figure}

\begin{figure}

    \begin{subfigure}{.3\columnwidth}    \centering

    \includegraphics[scale=0.25]{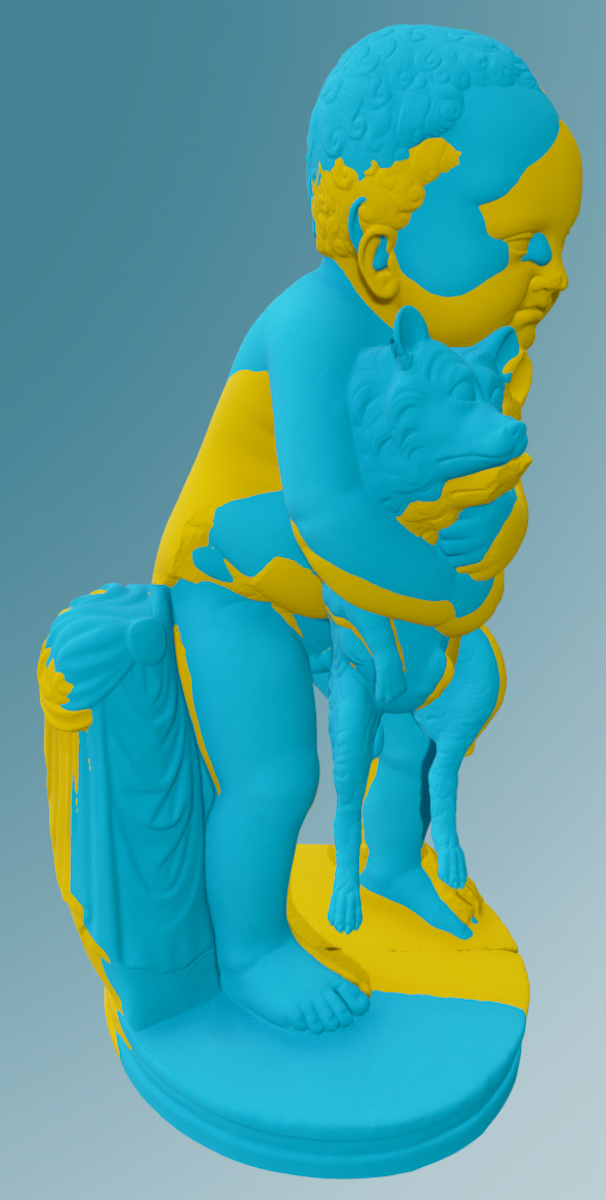}
    \end{subfigure}
    \begin{subfigure}{.3\columnwidth}    \centering
    \includegraphics[scale=0.25]{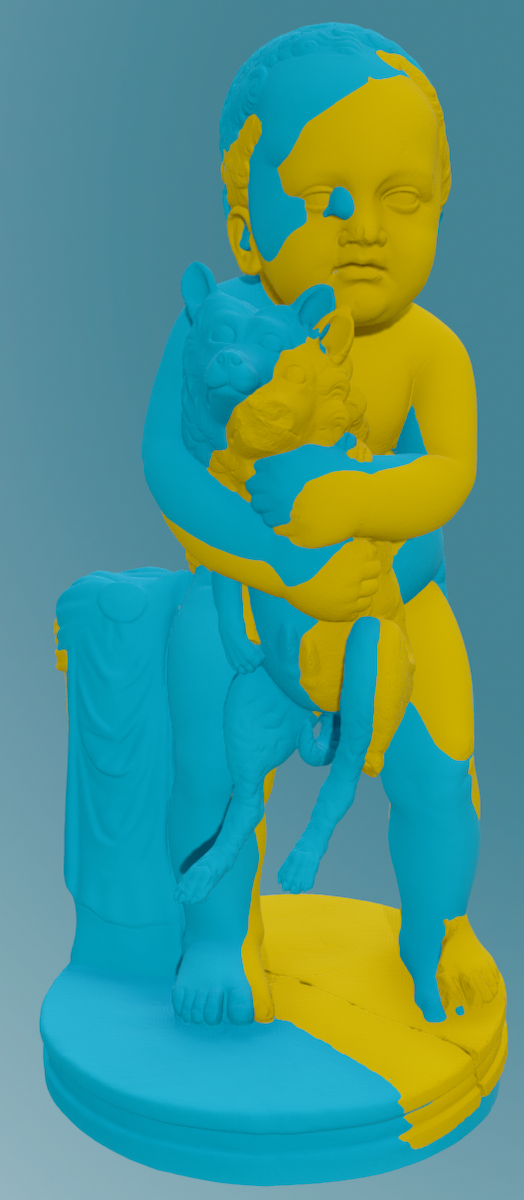}
    \end{subfigure}
    \begin{subfigure}{.3\columnwidth}    \centering
\includegraphics[scale=0.25]{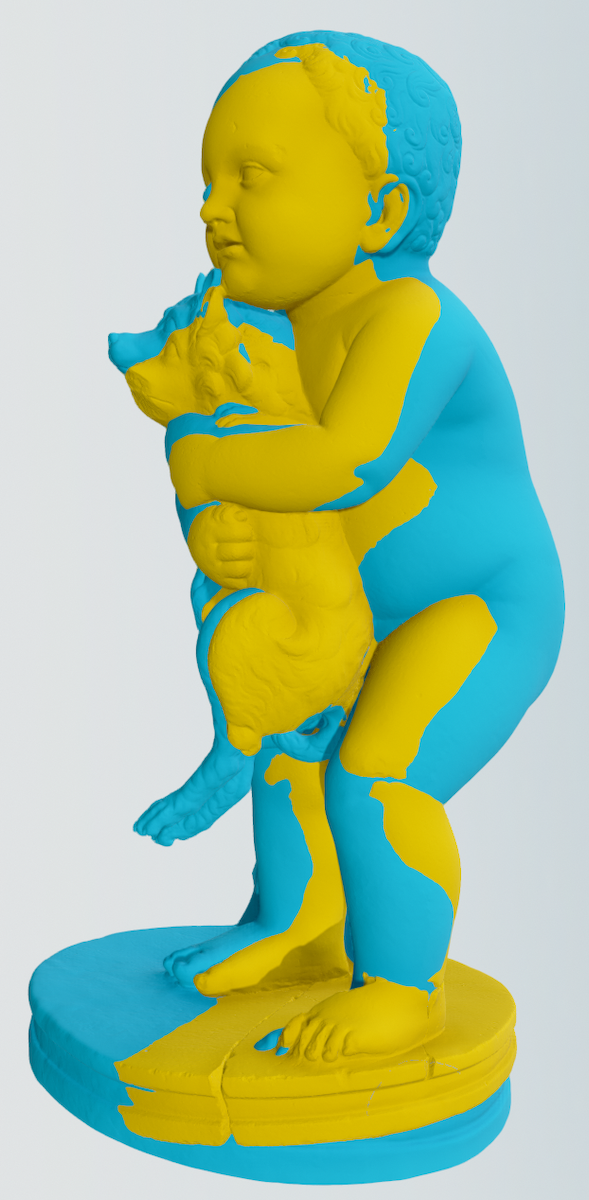}
    \end{subfigure}
    \caption{Testing RoAM: sample $400$ points from the broken statue mesh for point cloud $X$, apply a random linear map $L$ with $\mathrm{cond}\,L = 5.0$ to the restored statue mesh and sample another $400$ points from it for point cloud $Y$. Showing the  source and the image of the target under $L^{-1}_0\,L$. Frontal and side views.}\label{statue-400-400-5.0}
\end{figure}

In Figure~\ref{statue-broken-restored} (center) the ground truth of matching the damaged and restored sculptures is shown. In Figure~\ref{statue-600-600-3.0}--\ref{statue-400-400-5.0}, we show the matching of the broken sculpture (from which $X$ is sampled) to the restored one with a linear transform $L$ applied (from which $Y$ is sampled). In order not to produce a confusing image of the sculpture distorted by $L$, we instead show the image of the restored sculpture under $L^{-1}_0 L$ (where $L$ is the original linear transform, and $L_0$ is recovered by running RoAM) followed by a translation. The translation adjusts the barycenters of the two meshes.

Affine registration cannot recover the initial map as well as
rigid registration
in the presence of occlusion and noise: the affine group has more degrees of freedom, and thus it will necessarily register some bias produced by the noise and occlusion points. Note that in Figure~\ref{statue-600-600-3.0} with more sample points and a smaller condition number of the random linear map, we have a better match than in Figure~\ref{statue-400-400-5.0} where the number of sample points is reduced and the condition number of the random linear map is larger. 

The code and mesh data  are available on GitHub \cite{github:2023}.

\section{Comparison to other algorithms}
In this section, we compare RoAM to another algorithm, GrassGraf \cite{GrassGraph}, which appears to be the state--of--the--art  algorithm  to recover affine correspondences. 

We implemented  GrassGraph as described by its authors in \cite{GrassGraph} and applied it in the simplest case of two point clouds $X, Y \subset \mathbb{R}^{d \times n}$ where $X$ is the specimen cloud to be compared to $Y = L\, X\, S$ with $L \in \mathrm{GL}(d)$, $S \in \mathrm{Sym}(n)$.

The main finding is that GrassGraf loses precision very quickly if $Y$ is affected even by relatively small noise. The test point clouds  used are the unoccluded Caerbannog clouds \cite{caerbannog:2020}. We measured the values $\delta_L$ and $\delta_Y$, just as in the case of RoAM. These values are  enough to conclude that GrassGraph does not appear to recover $L$ robustly. 

\begin{figure}
    \begin{minipage}{.47\columnwidth}

        \centering
        \includegraphics[scale=0.35]{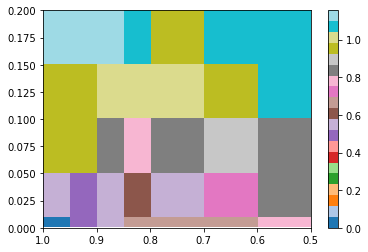}
        \caption{ ``Cow'' cloud for GrassGraph: $\delta_L$ as a function of $\lambda$ (discrepancy level, horizontal) and $\sigma$ (noise level, vertical) }
        \label{fig:grass_graph_cow_err_L}
   \end{minipage}
    \hspace{0.075cm}
\begin{minipage}{.47\columnwidth}
        \centering
        \includegraphics[scale=0.35]{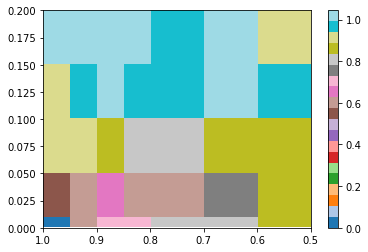}
        \caption{ ``Cow'' cloud for GrassGraph: $\delta_Y$ as a function of $\lambda$ (discrepancy level, horizontal) and $\sigma$ (noise level, vertical) }
        \label{fig:grass_graph_cow_err_Y}
        \end{minipage}
    \end{figure}

\begin{figure}
\begin{minipage}{.47\columnwidth}

        \centering
        \includegraphics[scale=0.35]{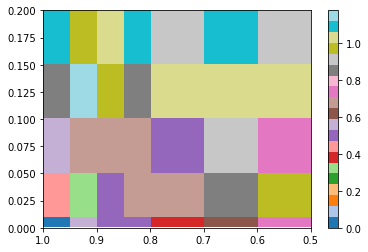}
        \caption{ ``Teapot'' cloud for GrassGraph: $\delta_L$ as a function of $\lambda$ (discrepancy level, horizontal) and $\sigma$ (noise level, vertical) }
        \label{fig:grass_graph_teapot_err_L}
        \end{minipage}
        \hspace{0.075cm}
\begin{minipage}{.47\columnwidth}

        \centering
        \includegraphics[scale=0.35]{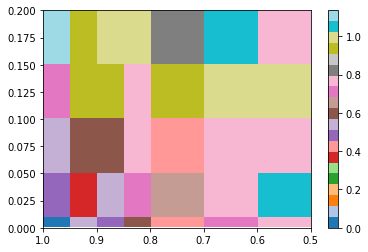}
        \caption{ ``Teapot'' cloud for GrassGraph: $\delta_Y$ as a function of $\lambda$ (discrepancy level, horizontal) and $\sigma$ (noise level, vertical) }
        \label{fig:grass_graph_teapot_err_Y}
        \end{minipage}
    \end{figure}

In Figures \ref{fig:grass_graph_teapot_err_L} -- \ref{fig:grass_graph_cow_err_Y}, we show the graphical depiction of our test results (as in the  case of RoAM) for GrassGraph with $3$ Laplacian eigenvectors used. The analogous dataset with test results for GrassGraph with $10$ Laplacian eigenvectors used is available on GitHub \cite{github:2023}. We do not reproduce it here as it shows only a marginal improvement. 

Also, some images of the test clouds are available on GitHub \cite{github:2023}: we do not reproduce them here,  note that the ``Teapot'' cloud shows slightly better robustness to noise and point discrepancy while, say, the ``Cow'' cloud already demonstrates a great difference between the recovered $L_0$ and the initial $L$ even for relatively small noise and point discrepancy. 

\section{Conclusions}
Feature matching is at the heart of many applications and
requires robust methods for recovery of correspondences and estimation of geometric transformations between domains.

This paper provides a robust and efficient algorithm to find correspondences between point clouds up to an affine transformation and gives bounds on the  algorithm's robustness in the presence of noise and point discrepancies. 

This paper also continues the line of research started in \cite{10155262}, and aims to generalize it from recovering orthogonal transformations (that are ``rigid'' in nature) to the case of much ``softer'' affine correspondences. 

\section{Acknowledgements}  This research is supported by the University of Austin (UATX) and by the Israel Science Foundation (ISF).

\bibliographystyle{elsarticle-num} 
\bibliography{refs}

\section{Appendix:  numerical tests}

~

    %
    %
    \begin{figure}
        \centering
        \includegraphics[scale=0.42]{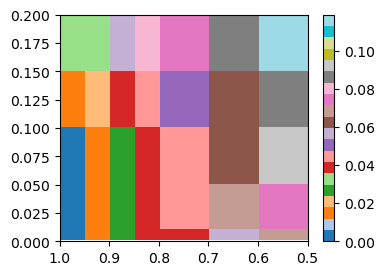}
        \caption{ ``Cow'' cloud: $\delta_Y$ as a function of $\lambda$ (discrepancy level, horizontal) and $\sigma$ (noise level, vertical) }
        \label{fig:cow_err_Y}
    \end{figure}
    \begin{figure}

    \centering
    \includegraphics[scale=0.0]{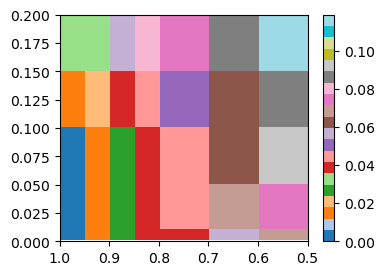}
    %
    %
\end{figure}


     \begin{figure}
        \centering
        \includegraphics[scale=0.0000]{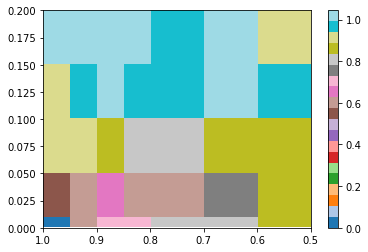}
    \end{figure}

    %
    %
    \begin{figure}
    \begin{minipage}{.47\columnwidth}

        \centering
        \includegraphics[scale=0.35]{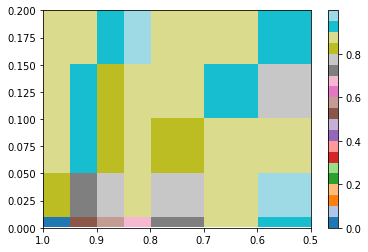}
        \caption{ ``Bunny'' cloud for GrassGraph: $\delta_L$ as a function of $\lambda$ (discrepancy level, horizontal) and $\sigma$ (noise level, vertical) }
        \label{fig:grass_graph_bunny_err_L}
    \end{minipage}
    \hspace{0.075cm}
\begin{minipage}{.47\columnwidth}
        \centering
        \includegraphics[scale=0.35]{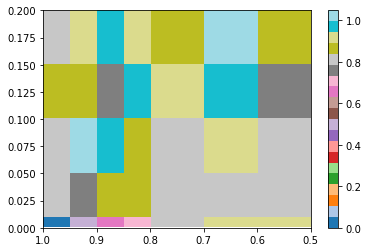}
        \caption{ ``Bunny'' cloud for GrassGraph: $\delta_Y$ as a function of $\lambda$ (discrepancy level, horizontal) and $\sigma$ (noise level, vertical) }
        \label{fig:grass_graph_bunny_err_Y}
        \end{minipage}
    \end{figure}

    \end{document}